\documentclass[conference]{IEEEtran}
\usepackage{svg}
\usepackage{amsmath} 
\usepackage{cite}
\ifCLASSINFOpdf
\else
\fi

\usepackage[hidelinks]{hyperref}

\usepackage{orcidlink}
\usepackage{listings}
\usepackage{tikz}

\newcommand\copyrighttext{%
  \footnotesize \textcopyright\>\the\year{} IEEE. Personal use of this material is permitted.  Permission from IEEE must be obtained for all other uses, in any current or future media, including reprinting/republishing this material for advertising or promotional purposes, creating new collective works, for resale or redistribution to servers or lists, or reuse of any copyrighted component of this work in other works.}

\newcommand\copyrightnotice{%
\begin{tikzpicture}[remember picture,overlay]
\node[anchor=south,yshift=20pt] at (current page.south) {\fbox{\parbox{\dimexpr0.82\textwidth-\fboxsep-\fboxrule\relax}{\copyrighttext}}};
\end{tikzpicture}%
}

\begin{document}

\title{SHACL-SKOS based \\ Knowledge Representation of Material Safety Data Sheet (SDS) for the Pharmaceutical Industry}

\author{\IEEEauthorblockN{Brian Lu\,\orcidlink{0009-0000-2253-3535}, Dennis Pham\,\orcidlink{0000-0001-5288-095X}}
\IEEEauthorblockA{Purdue University \\
West Lafayette, IN 47906, United States \\
contact@brianlu.me, dennis@dennispham.me}
\and
\IEEEauthorblockN{Ti-chiun Chang\,\orcidlink{0009-0005-7388-6169}, Michael Lovette\,\orcidlink{0000-0002-5747-7366}, Terri Bui\,\orcidlink{0009-0008-7637-9738}, Stephen Ma\,\orcidlink{0000-0002-0310-7562}}
\IEEEauthorblockA{MRL, Merck \& Co., Inc. \\
Rahway, NJ 07065, United States \\
\{ti-chiun.chang, michael.lovette, yen.bui, stephen.ma\}@merck.com}}


%Telephone: (800) 555--1212, Fax: (888) 555--1212}


\maketitle
\copyrightnotice

\begin{abstract}
%In order to meet patient needs, the drug development and manufacturing process is becoming more complex. With this complexity, comes the needs for data co-optimization between information systems. 
 We report the development of a knowledge representation and reasoning (KRR) system built on hybrid SHACL-SKOS ontologies for globally harmonized system (GHS) material Safety Data Sheets (SDS) to enhance chemical safety communication and regulatory compliance. SDS are comprehensive documents containing safety and handling information for chemical substances. Thus, they are an essential part of workplace safety and risk management. However, the vast number of Safety Data Sheets from multiple organizations, manufacturers, and suppliers that produce and distribute chemicals makes it challenging to centralize and access SDS documents through a single repository. To accomplish the underlying issues of data exchange related to chemical shipping and handling, we construct SDS related controlled vocabulary and conditions validated by SHACL, and knowledge systems of similar domains linked via SKOS. The resulting hybrid ontologies aim to provide standardized yet adaptable representations of SDS information, facilitating better data sharing, retrieval, and integration across various platforms. This paper outlines our SHACL-SKOS system architectural design and showcases our implementation for an industrial application streamlining the generation of a composite shipping cover sheet.
\end{abstract}


% \fi
\IEEEpeerreviewmaketitle

\section{Introduction} \label{intro}

Knowledge representation and reasoning (KRR) provides a framework for encoding knowledge in a format that artificial intelligence (AI) systems can utilize to reason, learn, and interact with the world in increasingly sophisticated ways. In 1993, the World Wide Web Consortium (W3C) was founded to develop a set of international standards, the modern beginnings of standardized KRR, for information exchange via the world wide web. One popular KRR model is a knowledge graph which utilizes graph-structured representation and database to store and operate on interconnected entities. In general, these standards created a cohesive set of KRRs that have allowed different systems and devices to communicate effectively -- fostering an open and interoperable web environment. A number of efforts have been taken (e.g., W3C standards \cite{w3c_standards}, NexIOM \cite{nexiom}, regulatory ontologies \cite{dumontier2014semanticscience, arp2015building, iso2011information}, OBO \cite{obofoundry}, etc.) to address general data exchange issues with unified ontologies acting as templates of universal KRRs, defining concepts, terms, and their relationship and properties. However, as the world wide web continues to develop and grow, so are the KRRs as they are increasingly adopted in industry-specific domains. In the case of the pharmaceutical sector, a number of KRRs have already been developed which focus on drug-target identification, drug discovery, drug development\cite{lin2017drug, visser2011bao, griffith2013gpcr, wishart2006drugbank, hewett2017pharmgkb}, and health outcomes \cite{cella2010patient, smith2007obo, hripcsak2015observational, hanna2022chronic, michie2017human}.

 Recently, there has been growing interest in integrating systems related to pharmaceutical manufacturing, quality control, and regulatory compliance to more effectively address concerns related to the shipping and handling process. % any reference to justify the "growing interest"?
 This necessitates increased interface between information objects to ``connect the dots'' across different information domains (i.e., Enterprise Organization, Modeling \& Simulations, Process Development, Warehousing \& Procurement, Project Timelines, etc.) -- all of which have their own internal domain-specific data models and vocabularies. However, these systems often rely on individually defined ontologies, which upon integration, are faced with a ``terminology challenge'' as data exchange between systems becomes more complex.   Recent alliances, such as Pistoia \cite{pistoia_alliance} and Allotrope \cite{Millecam2021}, were formed to support the integration of information gathering, management, and model development of pharmaceutically relevant information in hopes of overcoming both the interoperability and terminology challenge associated with KRRs. In most cases, pharmaceutically relevant ontologies trend towards alignment with industrial standards for product and process development \cite{VENKATASUBRAMANIAN20061482, engproc2021009038, raebel2013standardizing, MORBACH2007147, mann2023susie}. 
 
 Unfortunately, user adoption of these individually aligned ontologies have been limited due to complexity and difficulty in implementation. For example, ontologies based on basic formal ontology (BFO) \cite{Arp2015-bk} encourage modeling vocabularies as classes, which are subclassed from other BFO classes, making them unwieldy and challenging to understand. As BFO expresses data in the form of overly faceted classes, reasoning at a certain higher-level requires significant and tedious introspection into the class structure. Specifically limiting the number of sub-classes to those that are structurally relevant allows implementers and users of the ontology to better understand how the ontology classes show a specific ``view'' of the data on one abstraction level. Taxonomy construction, to support fine-grained vocabularies, can then be offloaded to a knowledge organization system that is fit for the task, such as Simple Knowledge Organization System (SKOS), which is our preferred choice. By separating these concepts, it becomes easier to construct and integrate the ontologies operationally. Following such design principles, ontologies such as Quantities, Units, Dimensions, and Types (QUDT) \cite{qudt}, Units of Measure (UOM) \cite{uom}, and Medical Subject Headings (MeSH) \cite{mesh} have seen more success as they utilize SKOS framework to create domain specific taxonomies (hierarchical classification system with structured relationships between concepts) rather than overly intertwined ``top-level'' organization of concepts and terminologies, facilitating easier processing with software. 
 
To further elaborate, we utilize SHACL ``shapes'' (container-like objects) to define the structure and constraints of domain-specific data, as each shape can reference multiple SKOS ``taxonomies'' (controlled vocabularies and hierarchical organization of concepts). This approach enhances flexibility and re-usability for modeling complex systems. As such, application of these frameworks allows for modeling data with constraints and makes it easier to specify conditions and ensure data validity.

 \begin{figure}[ht]
    \centering
    \includesvg[width=2in]{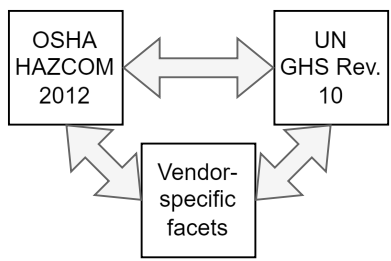}
    \caption{Minor vocabulary differences can introduce friction}
    \label{fig:ieee6}
\end{figure}

 In our specific pharmaceutical use-case, a seemingly straightforward process of creating a comprehensive packaging cover sheet, which includes material classifications and handling procedures,  requires the integration of discrete SDS across multiple internal and external vendor systems. As illustrated in Fig.\ref{fig:ieee6}, while the United Nations (UN) Globally Harmonized Systems (GHS) \cite{ghs} standardizes most language specific to the communication of hazard and safety information for substances and compounds, there is a need to cross reference to the Occupational Safety and Health Administration (OSHA) Hazard Communication (HazCom) standards\footnote{https://www.osha.gov/hazcom/HCS-Final-RegText}. Therefore, SDS in the wild have to adhere to older standards, are subject to vendor interpretation, or need to fulfill local regulatory requirements, resulting in difficulties to streamline cover sheet generation. In these cases, usage of SKOS enables multiple taxonomies of various standards to be applied to the same shape, realizing the benefits of integrating potentially heterogeneous vocabularies.

Here we present the development of a SHACL-SKOS data model to create ontologies that enables users to navigate complex SDS datasets intuitively. This framework aims to address the challenges of data integration and terminology inconsistencies in pharmaceutical development processes by leveraging simple and modular ontologies within a modern data platform architecture. The separation of (SHACL) shapes and (SKOS) taxonomies is a key aspect to this approach which allows shapes to reference multiple taxonomies and provides different views on shared concepts, facilitating better data sharing, retrieval, and integration across various platforms and stakeholders. By establishing a SHACL-SKOS framework for representing SDS information, we aim to improve chemical safety communication, enhance regulatory compliance efforts, and promote a culture of safety across industries. We report its design, explore its components in detail, and discuss its potential impact on the knowledge management of material safety.

%SKOS is an RDF standard designed to encompose thesauri, schema, and taxonomies and facilitate easy publication and data linkages. The SHACL component further supports hierarchical structuring, making it easier to identify and classify information via constraints of the contents structure and meaning of a KRR -- thereby enabling domain specific technical validation and constraints on a representation.

\section{Methods}

In this section, we outline our strategy for ontology construction through SHACL-SKOS concepts, workflow design decisions, and our current implementation of the SHACL-SKOS system for SDS data extraction and summarization.  %\\\\\\\\\\\\\\\\begin{figure}[h]
%    \\\\\\\\\\\\\\\\centering
%    \\\\\\\\\\\\\\\\includegraphics[width=3in]{figures/cover-sheet-redacted.png}
%    \\\\\\\\\\\\\\\\caption{Cover Sheet Generation}
%    \\\\\\\\\\\\\\\\label{fig:coversheet}
%\\\\\\\\\\\\\\\\end{figure}

\subsection{Ontology Modeling Strategy}
Our SHACL-SKOS system, referred to as DeepPharmGraph (DPG), adopts a graph-structured data model for potentially coupled (deep-learning based) inference and appears as the prefix of modules in Figs.\ref{fig:ieee5}-\ref{fig:ieee4}. It simplifies taxonomy construction using SKOS, facilitates integration of different vocabularies, and offers standardized terms for referencing by multiple shapes. These modules were constructed for the SHACL-SKOS system and are not a part of any public ontologies/taxonomies.  Additionally, our strategy, as elaborated below, supports multiple views on shared taxonomic concepts, enabling diverse models while maintaining consistency and reducing redundancy. 

\begin{itemize}
\item{\textbf{Separation of Shapes and Taxonomies:}} Recognizing that shapes and taxonomies serve different purposes, we separate them to enhance modularity and reusability. Shapes define the structure and constraints of data, while taxonomies provide the controlled vocabularies and hierarchical organization of concepts. By decoupling these components, shapes can reference multiple taxonomies, and taxonomies can be applied across different shapes. 
\item{\textbf{Simplified Taxonomy Construction:}} We utilize SKOS to create taxonomies that organize concepts in a hierarchical manner. This approach makes taxonomy construction straightforward and facilitates the integration of different vocabularies. Taxonomies like DPG-GHS (aligned with GHS SDS headings and hazard classifications) and DPG-ISA-88 (describing ISA-88 terms) provide standardized terms that can be referenced by multiple shapes. 
\item{\textbf{Multiple Views on Shared Taxa:}} Shapes represent different ``views'' (or perspectives) on the same set of taxonomic concepts. For example, the DPG-DoC shape models document structures, while the DPG-SafeD shape models safety data. Both shapes can reference the same taxonomy (e.g. DPG-GHS), enabling consistent interpretation and reducing redundancy. 
\end{itemize}

\subsection{Shapes and Taxonomies}

\begin{figure}[ht]
    \centering
    \includesvg[width=2.4in]{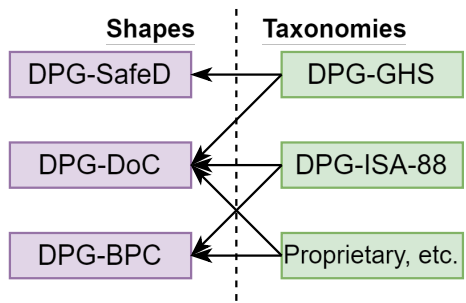}
    \caption{Shape graphs and their related taxonomies}
    \label{fig:ieee5}
\end{figure}

% As the SHACL-SKOS methodology inherently decouples shapes and taxonomy components, shapes can then reference multiple taxonomies, and taxonomies can be applied across different shapes. % repeated
 To simplify taxonomy construction, SKOS organizes concepts in a hierarchical manner. Scoping of allowed taxa and extended properties is facilitated via subclassing of \texttt{skos:Concept} (user-defined versus inherited within the broader SKOS concept scheme). This approach makes taxonomy construction straightforward and facilitates the integration of different vocabularies. 
 %repeated%%For this SDS use case, taxonomies like DPG-GHS (aligned with GHS SDS headings and hazard classifications) and DPG-ISA-88 (describing ISA-88 terms) were constructed to provide standardized terms that can be referenced by multiple shapes. 

To accurately capture the relationships between different concepts within SDS documents, ensure semantic clarity, and enhance data interoperability, we employed SKOS, and additionally, SKOS eXtension for Labels (SKOS-XL) for application-specific extensions \cite{w3SKOSXL}: GHS datasheet sections (for use in documents) or GHS hazards (for use in classification) are represented as SKOS Concepts. The SHACL shapes for SKOS are derived from those used by SkoHub \cite{skohub}. SKOS-XL labels are used as additional data when performing document understanding heuristics or rendering user interfaces.

Usage of SKOS-XL is envisioned as a means to accelerate construction of ontology by performing the heavy-lifting of taxonomy and controlled vocabulary construction (See Fig. \ref{fig:ieee5}) by forming the basis of a knowledge organization system. In this development, ontology files reference the following shapes and taxonomies.
\\\\
\noindent\textbf{Shapes}

\noindent\subsubsection*{DPG-DoC (Document Components)} Defines the structure of documents, such as sections, headings, and containers. It provides a schema for representing document components, allowing for consistent parsing and interpretation of various document formats. DPG-DoC has re-written parts of SPAR Pattern Ontology \cite{sparontologiesPatternOntology}, DoCO \cite{sparDoCO}, and FaBiO \cite{PERONI201233} in the form of SHACL shapes and a SKOS document. The SPAR Pattern Ontology is taken relatively directly, translated to SHACL, representing the structural semantics of a document. The SPAR DoCO ontology is broken down into a SKOS-XL taxonomy as marker concepts for use in shapes. This also opens up opportunity to extend the taxonomy to create more semantically fine-grained markers, such as those for GHS SDS document headings, as opposed to generic headings.

\begin{figure}[ht]
    \centering
    \includesvg[width=2.90in]{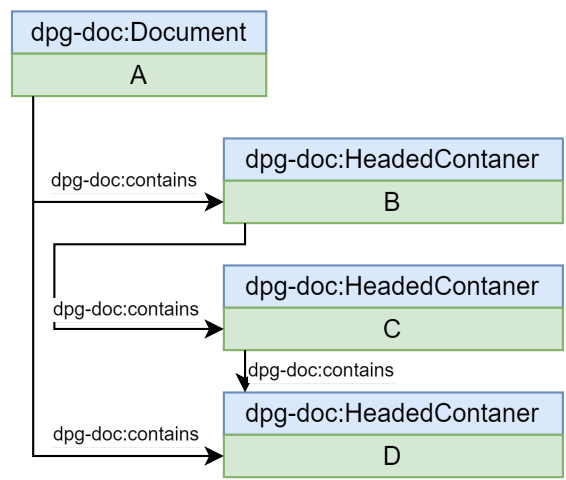}
    \caption{Sample Document Components Shape}
    \label{fig:ieee1}
\end{figure}

A document consists primarily of hierarchical containers, but can also represent other shapes such as floats and header/footer metadata (See Fig. \ref{fig:ieee1}).

\begin{figure}[ht]
    \centering
    \includesvg[width=2.50in]{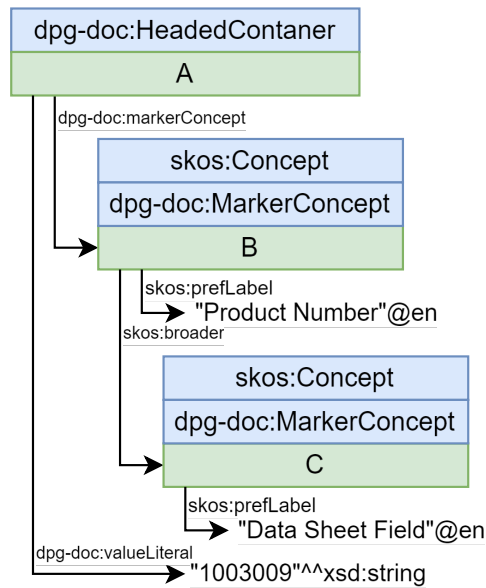}
    \caption{Document components shape demonstrating support with MarkerConcept objects}
    \label{fig:ieee2}
\end{figure}

Containers may contain literals and/or concept references. Literals are useful for data values that are not part of a taxonomy, such as numerical values. Otherwise, concepts are preferred, since they can be reused across multiple data and shape graphs (See Fig. \ref{fig:ieee2}).

\noindent\subsubsection*{DPG-SafeD (Safety Data)} Models structured representations of safety data, including hazard classifications and safety measures. It defines how safety-related information is organized and constrained within the data.

\begin{figure}[ht]
    \centering
    \includesvg[width=3.1in]{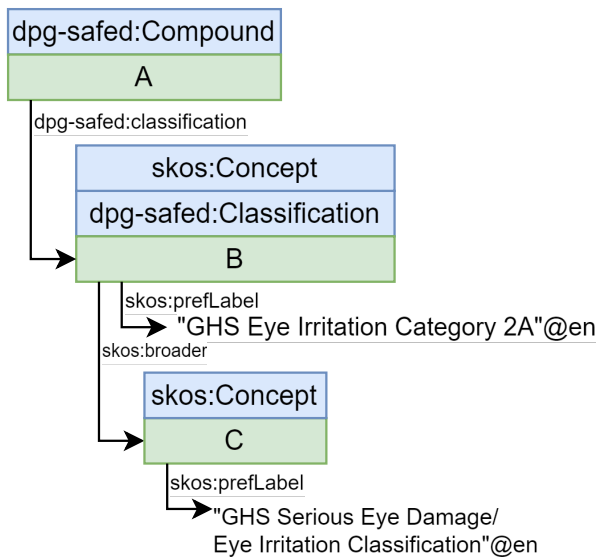}
    \caption{Sample Safety Data Shape demonstrating compound hazard classification }
    \label{fig:ieee3}
\end{figure}

SafeD is described by shapes with the purpose of assigning classifications (which are subclasses of SKOS Concepts). For example, a SafeD compound can have a ``GHS Eye Irritation Category 2A'' classification assigned to it, since the SKOS Concept for it is also a SafeD Classification (See Fig. \ref{fig:ieee3}). This classification concept can also be integrated into a broader taxonomy independent of the SafeD shapes.

\begin{figure}[ht]
    \centering
    \includesvg[width=2.6in]{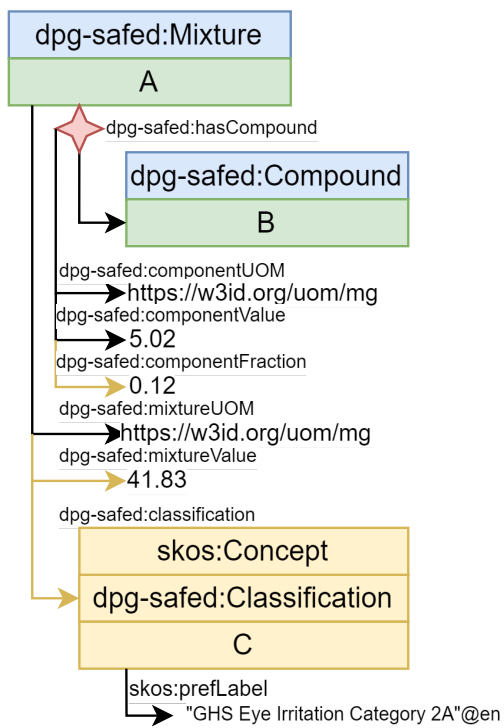}
    \caption{Safety Data Shape demonstrating SHACL-AF rulesets hazard inference enablement for Mixtures of Compounds}
    \label{fig:ieee4}
\end{figure}

Adding Mixture hazard classification into the mix provides opportunities to utilize the capacity to express inference rules via SHACL-AF. In this case, suppose a Mixture can be automatically classified as a GHS Eye Irritation Category 2A because one of its constituent ingredients exceeds a 10\% concentration, and is also classified as 2A (See Fig. \ref{fig:ieee4}). 

\noindent\subsubsection*{DPG-BPC (Batch Process Control)} Represents batch process control structures, aligning with industry standards like ANSI/ISA-88. It models procedural and physical aspects of batch processes. This shape intends to serve as the semantic view into the data represented in batch process records, and aims to ease the composition of data from multiple of such records.
\\\\
\noindent\textbf{Taxonomies}
\noindent\subsubsection*{DPG-GHSrev10} A taxonomy aligned with the Globally Harmonized System (GHS) for classifying and labeling chemicals. It includes concepts for SDS headings, hazard classification categories, and specific hazard classes. This taxonomy can be integrated into both DPG-DoC (as document headings and values) and DPG-SafeD (as safety classifications).

\noindent\subsubsection*{DPG-ISA-88} A taxonomy that describes terms from the ISA-88 standard for batch process control. It provides standardized terminology that can be applied to both DPG-DoC and DPG-BPC shapes.

\noindent\subsubsection*{Potential Taxonomies} Additional hazard taxonomies, which cover the 2012 OSHA HazCom specification, proprietary company batch control processes, and Work Instructions stored in a GMP document store, can be created and used in the same manner as these other taxonomies. The modular architecture of the system allows additional terms to be easy to integrate.

\subsubsection*{Integration of Shapes and Taxonomies}
Using our use case, we elaborate the advantages of separating shapes and taxonomies for greater flexibility as shapes can reference multiple taxonomies, enabling them to incorporate concepts from various domains:
\begin{itemize}
    \item DPG-DoC can utilize taxonomies like DPG-GHS to represent SDS headings within document structures, and DPG-ISA-88 to model process documentation.
    \item DPG-SafeD references DPG-GHS to model safety classifications and hazard information within safety data.
    \item DPG-BPC integrates DPG-ISA-88 and proprietary taxonomies to represent batch process control procedures, such as tablet compression processes.
\end{itemize}
This approach reduces the complexity of ETL processes, as data can be transformed more easily when utilizing a common vocabulary.

\subsection{Platform and Workflow Design}
\noindent\textbf{Industry-Standard Data Platform:} A modern data platform that leverages industry-standard technologies was deployed for this use-case. The platform is built around a service-oriented architecture (SOA) provisioned using Infrastructure as Code (IaC) on Kubernetes clusters in a hybrid cloud environment. This setup ensures scalability, flexibility, and integration with existing infrastructure.
\\\\
\noindent\textbf{Document-Centric ETL Workflows:} The platform focuses on extracting, transforming, and loading (ETL) documents, whether digitally rendered or scanned. Techniques such as Named Entity Recognition (NER), Optical Character Recognition (OCR), and Vision-Language Models (VLM) are used to process documents. By describing extracted data using our standard taxonomies, we facilitate easier transformation and integration into different views.
\\\\
\noindent\textbf{Workflow Orchestration with Apache Airflow:} Apache Airflow  was utilized to orchestrate workflows and enable data-driven scheduling. Airflow allows for the separation of dependencies using virtual environments and supports the use of Kubernetes operators for scalable workloads.
\\\\
\noindent\textbf{Data Storage and Access:} The triplestore, Oxigraph, serves as the system's data store due to its support for read replicas and ease of embedding in applications and workflows. This enables efficient querying and data retrieval using SPARQL.
\\\\
\noindent\textbf{Identity and Access Management:} Keycloak provides identity and access management, integrating with internal identity providers and supporting role-based access control. This ensures secure access to data and services.

\subsection{Workflow Overview}
\noindent The overall Crucible workflow involves several key steps:

\begin{itemize}
    \item \textbf{Document Ingestion:} Users upload documents such as SDS, batch records, or work instructions. The system ingests these documents for processing via web interface.
    \item \textbf{Data Extraction and Annotation:} The system uses NER, OCR, and VLM methods to extract relevant data from the documents. Extracted data is annotated using the appropriate taxonomies (e.g., DPG-GHS, DPG-ISA-88) and validated against the corresponding shapes (e.g., DPG-DoC, DPG-SafeD).
    \item \textbf{Data Integration and Storage:} Annotated data is stored in the triplestore, enabling complex queries and integration with other data sources.
    \item \textbf{Data Transformation:} Standardized data can be transformed into different views or formats as needed. For example, safety data extracted from SDS can be integrated into batch process records or used to generate composite packing sheets.
    \item \textbf{User Interaction:} Users interact with the system through custom interfaces. They can perform tasks such as generating reports, analyzing process variables, or querying hazard classifications.
    \item \textbf{Interoperability and Data Exchange:} The use of standardized taxonomies and shapes facilitates data exchange with other systems, reducing the need for complex ETL transformations.
\end{itemize}

\section{Results}

% \begin{figure}[h]
%     \centering
%     \includesvg[width=3in]{figures/ieee7-overview.drawio.svg}
%     \caption{Multiple shapes with a common taxonomy enables simplified transformations from one to the other. The chemical safety shape shown here includes a rule set to infer a label element used in the cover sheet.}
%     \label{fig:transformoverview}
% \end{figure}

\noindent As a practical example, the conventional method of generating a ``Test Product'' composite packing sheet for shipping materials is quite manual. Existing SDS platforms utilize fairly rigid search mechanisms that operate on a set of fixed string search and ``drill-down'' widgets which require users to apply a number of filters to navigate the search results \(S^{E}\) for compound \(E\). \(S^{E}\) is further distinguished by Manufacturer \(i\), Language \(j\), and Revision Date \(k\). For easier conceptualization, this process is denoted by the following ``equation'': 

\begin{figure*}[hbt!]
    \centering
    \includegraphics[width=6in]{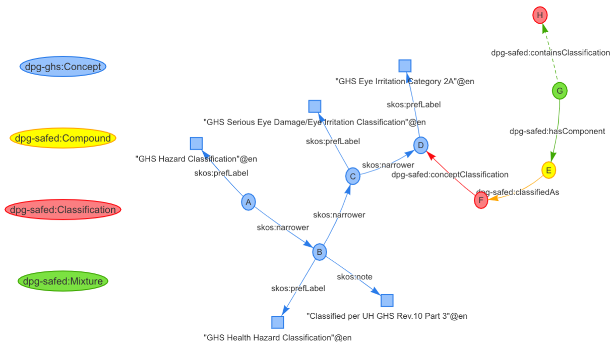}
    \caption{Hazard Statements:  Truncated hazard SDS network visualization traversing multiple taxonomies}
    \label{fig:network}
    
\end{figure*}

\begin{figure}[ht]
\centering
\begin{lstlisting}[language=SPARQL,basicstyle=\ttfamily\small]
SELECT ?hazard ?prefLabel ?labelDisplay
WHERE {
  <E> safed:classification ?hazard .
  ?hazard skos:prefLabel ?prefLabel .
  ?hazard safed:labelDisplay ?labelDisplay .
  FILTER (lang(?prefLabel) = "en")
}
\end{lstlisting}
\caption{Example of a query an application may use to retrieve the information necessary to display the hazard classifications of a compound.}
\label{fig:sparql}
\end{figure}

\begin{equation}\label{eq:1}
S^{E} =  \sum S^{E}_{i,j,k}
\end{equation}

\begin{figure}[ht]
    \centering
    \includegraphics[width=3.4in]{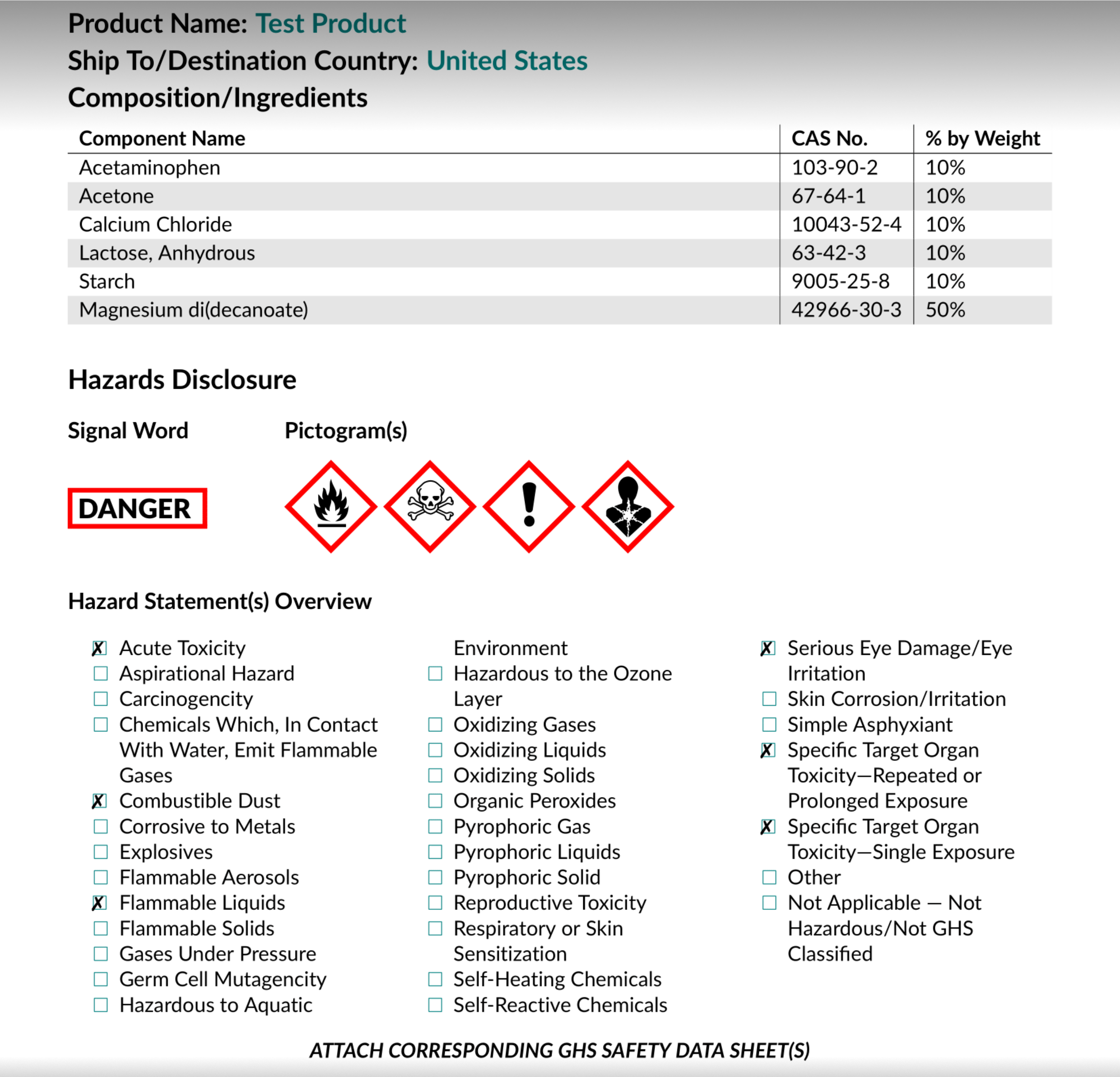}
    \caption{Example cover sheet generated by Crucible}
    \label{fig:coversheet}
\end{figure}

\noindent For multiple compounds $\overrightarrow{N}=\{E_1, E_2, ...\}$ contained in a product, search operations to obtain the set of all related SDS \(S^{\overrightarrow{N}}\) would then need to be manually performed (search and download) for every compound $E_i$ where $ i=1,2,...$, as accounted for in a composite shipping and handling cover sheet (e.g. ``Composition/Ingredients'' in Fig. \ref{fig:coversheet}): 
 
 \begin{equation}\label{eq:2}
S^{\overrightarrow{N}} = \sum_{i} S^{E_i}
 \end{equation}

\noindent After obtaining $S^{\overrightarrow{N}}$, users have to compile a comprehensive set of Hazards \(H^{\overrightarrow{N}}\) from hazard statements $D^{E_i}_{P_j}$ listed in different sections $P_j$ in each $S^{E_i}$ of $S^{\overrightarrow{N}}$. To make it explicit for the required manual compilation, we write: 

\begin{equation}\label{eq:3}
H^{\overrightarrow{N}} = \sum_{i}\sum_{j}D^{E_i}_{P_j}
\end{equation}

\noindent Once obtained, hazards of $H^{\overrightarrow{N}}$ are validated against a list of general hazards statements \(H^{gen}\) to get the final set of hazard statements $\{D\}$ to be disclosed on the comprehensive cover sheet (see, for example, ``Hazards Disclosure" and ``Hazard Statement(s) Overview" in Fig. \ref{fig:coversheet}) as: 

\begin{equation}\label{eq:4}
H^{\overrightarrow{N}} \cap H^{gen} = \{ D : D\in H^{\overrightarrow{N}}  \text{ and } D \in H^{gen}\}
\end{equation}

\begin{figure}[ht]
    \centering
    \includegraphics[width=3in]{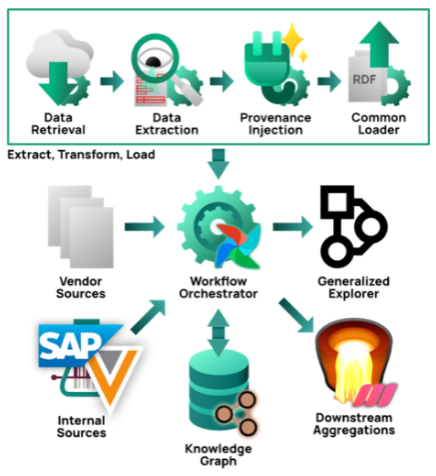}
    \caption{Data Platform Services:  This platform automates manual tasks, reduces errors, and ensures consistency in safety information.}
    \label{fig:resultsprocess}
\end{figure}

\noindent This manual process has since been automated through the development of an internal system code-named ``Crucible'' (see the modules and workflow in Fig. \ref{fig:resultsprocess}) that utilizes our SKOS-SHACL methodology. 

Using Crucible, users can upload SDS to a graphical user interface (GUI) or select from pre-existing SDS documents. Uploaded SDS are processed via a hosted workflow orchestrator that extracts the Safety Data Sheet information conformant to the SHACL-SKOS models. Specifically, Hazard Pictograms and Hazard Statements are extracted and annotated using DPG-GHS taxonomy and validated against SafeD shape. The extracted information from multiple SDS, is integrated into a downstream knowledge graph -- allowing for aggregation and analysis. Finally, a composite packing sheet is generated by querying the associated labels for multiple SDS. With the returned triples, a table look-up operation is performed to link and return the `Hazard Disclosure' components of SDS into the required composite packing SDS sheet format (PDF) as a summary. 

Essentially, Crucible has automated the manual processes described in Eqs. \ref{eq:1}--\ref{eq:4} and is able to obtain \(H^{\overrightarrow{N}}\) from an internal knowledge graph. Once the user provides the necessary information (SDS selection), iteration is simply utilized to generate the appropriate SPARQL query for Eq. \ref{eq:3} for each compound \(E\) to obtain a Hazard Statement from various sections. Figure \ref{fig:network} illustrates a truncated network visualization of \(H^{\overrightarrow{N}}\) generated from querying the hazard statements for `Acetomenophin 400' \cite{SigmaAldrich_1003009}. Rather than performing the manual operations, we are able to recall \(H^{\overrightarrow{N}}\) with a simple SPARQL query (see Fig. \ref{fig:sparql}) across multiple vendors and through multiple SDS in the English language without relying on complex language by using multiple taxonomies. Using Crucible, we were able to reduce the manual processing time of generating a composite shipping cover sheet, as shown in Fig. \ref{fig:coversheet}, from hours to minutes.

% \begin{figure*}[hbt!]
%     \centering
%     \includegraphics[width=6in]{figures/network2.png}
%     \caption{Hazard Statements:  Truncated hazard SDS network visualization traversing multiple taxonomies}
%     \label{fig:network}
    
% \end{figure*}

% \begin{figure}[h]
% \centering
% \begin{lstlisting}[language=SPARQL]
% SELECT ?hazard ?prefLabel ?labelDisplay
% WHERE {
%   <E> safed:classification ?hazard .
%   ?hazard skos:prefLabel ?prefLabel .
%   ?hazard safed:labelDisplay ?labelDisplay .
%   FILTER (lang(?prefLabel) = "en")
% }
% \end{lstlisting}
% \caption{Example of a query an application may use to retrieve the information necessary to display the hazard classifications of a compound.}
% \label{fig:sparql}
% \end{figure}

Additional benefits to this process are focused on enhanced user experience in which users are able to generate the summary report for shipping and handling information by navigating through a catalog of processed SDS compounds (using fuzzy filtering) and adding them to a ``shopping-cart''. This process is a significant departure from the traditional methods of filtering through drop-down widgets and fixed string search UX-elements for the SDS-specific platforms.

%\begin{itemize}
%    \item \textbf{Document Ingestion:} SDS sheets are uploaded and ingested into the system.
%    \item \textbf{Data Extraction:} Safety data is extracted and annotated using DPG-GHS taxonomy and validated against DPG-SafeD shape.
%    \item \textbf{Data Integration}: Extracted safety data is integrated into the knowledge graph, allowing for aggregation and analysis.
%    \item \textbf{Report Generation:} A composite packing sheet is generated by transforming the integrated data into the required format, utilizing standardized terms and classifications.
%    \item \textbf{Benefits:} This process automates manual tasks, reduces errors, and ensures consistency in safety information.
%\end{itemize} 

\section{Conclusion}
% TODO: review
%% Ti-chiun and Stephen Here
There has been significant effort put into the construction of ontologies and KRR in the pharmaceutical industry, as discussed in Section \ref{intro}. However, the adoption of the KRR models into operational systems has been slow. Among many challenges are the agreement on vocabulary, the complexity of integrating knowledge of multiple domains, and the lack of an adaptive, scalable, and flexible SHACL-SKOS KRR system architecture. Our implementation, with its flexibility and attentiveness, shows a promising pathway to overcome many of these challenges. By separating shapes and taxonomies and allowing shapes to reference multiple taxonomies, we achieve:

\begin{itemize}
    \item \textbf{Modularity:} Components can be developed, maintained, and updated independently, enhancing flexibility.
    \item \textbf{Reusability:} Taxonomies can be reused across different shapes and applications, reducing duplication of effort.
    \item \textbf{Interoperability:} Standardized vocabularies and structures enable seamless data exchange between systems.
    \item \textbf{Reduced Complexity:} Simplifying the modeling approach lowers the complexity of ETL processes and data transformations.
    \item \textbf{Scalability:} The architecture supports scaling to accommodate growing data volumes and additional domains.
\end{itemize}
In particular, SKOS(-XL) offers the capability to incorporate alternative expressions (in the form of languages, terms, related concepts) and SHACL can enforce data quality assurance. Moreover, our emphasis on interoperability and incorporation of multiple ontologies demonstrates the potential to expand to related workflows in the pharmaceutical development, although our current use case is limited to KRR of pharmaceutically relevant material SDS.

\bibliographystyle{IEEEtran}
\bibliography{ref}

\begin{thebibliography}{10}
\providecommand{\url}[1]{#1}
\csname url@samestyle\endcsname
\providecommand{\newblock}{\relax}
\providecommand{\bibinfo}[2]{#2}
\providecommand{\BIBentrySTDinterwordspacing}{\spaceskip=0pt\relax}
\providecommand{\BIBentryALTinterwordstretchfactor}{4}
\providecommand{\BIBentryALTinterwordspacing}{\spaceskip=\fontdimen2\font plus
\BIBentryALTinterwordstretchfactor\fontdimen3\font minus \fontdimen4\font\relax}
\providecommand{\BIBforeignlanguage}[2]{{%
\expandafter\ifx\csname l@#1\endcsname\relax
\typeout{** WARNING: IEEEtran.bst: No hyphenation pattern has been}%
\typeout{** loaded for the language `#1'. Using the pattern for}%
\typeout{** the default language instead.}%
\else
\language=\csname l@#1\endcsname
\fi
#2}}
\providecommand{\BIBdecl}{\relax}
\BIBdecl

\bibitem{w3c_standards}
\BIBentryALTinterwordspacing
{World Wide Web Consortium}, ``W3c standards,'' 2024, accessed: 2024-10-09. [Online]. Available: \url{https://www.w3.org/standards/}
\BIBentrySTDinterwordspacing

\bibitem{nexiom}
R.~Hodgson, TopQuadrant, and N.~N.~O. Lead, ``{N}{E}x{I}{O}{M}, the {N}{A}{S}{A} {C}onstellation {P}rogram {O}ntologies,'' \url{https://step.nasa.gov/pde2009/slides/20090506145822/PDE2009-NExIOM-TQ_v2.0-aRH-sFINAL.pdf}, 2009.

\bibitem{dumontier2014semanticscience}
M.~Dumontier, C.~J. Baker, J.~Baran, A.~Callahan, L.~Chepelev, J.~Cruz-Toledo, N.~R. Del~Rio, G.~Duck, L.~I. Furlong, N.~Keath \emph{et~al.}, ``The semanticscience integrated ontology (sio) for biomedical research and knowledge discovery,'' \emph{Journal of biomedical semantics}, vol.~5, no.~1, pp. 1--11, 2014.

\bibitem{arp2015building}
\BIBentryALTinterwordspacing
R.~Arp, B.~Smith, and A.~D. Spear, \emph{Building ontologies with basic formal ontology}.\hskip 1em plus 0.5em minus 0.4em\relax MIT Press, 2015. [Online]. Available: \url{https://mitpress.mit.edu/books/building-ontologies-basic-formal-ontology}
\BIBentrySTDinterwordspacing

\bibitem{iso2011information}
\BIBentryALTinterwordspacing
{International Organization for Standardization}, \emph{Information and documentation-Bibliographic references and citations}, ISO Std. ISO 690:2010, 2011. [Online]. Available: \url{https://www.iso.org/standard/43320.html}
\BIBentrySTDinterwordspacing

\bibitem{obofoundry}
``{O}{B}{O} {F}oundry,'' \url{https://obofoundry.org/principles/fp-000-summary.html}.

\bibitem{lin2017drug}
Y.~Lin, S.~Mehta, M.~Uhlén, A.~Mardinoglu, and J.~Nielsen, ``Drug target ontology to classify and integrate drug discovery data,'' \emph{Journal of Biomedical Semantics}, vol.~8, no.~1, pp. 1--11, 2017.

\bibitem{visser2011bao}
U.~Visser, S.~Abeyruwan, U.~Vempati, R.~P. Smith, V.~Lemmon, and S.~C. Schürer, ``Bioassay ontology (bao): a semantic description of bioassays and high-throughput screening results,'' \emph{BMC Bioinformatics}, vol.~12, no.~1, pp. 1--12, 2011.

\bibitem{griffith2013gpcr}
M.~Griffith, O.~L. Griffith, A.~C. Coffman, J.~V. Weible, J.~F. McMichael, N.~C. Spies, J.~Koval, I.~Das, M.~B. Callaway, J.~M. Eldred \emph{et~al.}, ``Gpcr ontology: development and application of a g protein-coupled receptor pharmacology knowledge framework,'' \emph{Bioinformatics}, vol.~29, no.~24, pp. 3211--3219, 2013.

\bibitem{wishart2006drugbank}
D.~S. Wishart, C.~Knox, A.~C. Guo, S.~Shrivastava, M.~Hassanali, P.~Stothard, Z.~Chang, and J.~Woolsey, ``Drugbank: a comprehensive resource for in silico drug discovery and exploration,'' \emph{Nucleic acids research}, vol.~34, no. suppl\_1, pp. D668--D672, 2006.

\bibitem{hewett2017pharmgkb}
M.~Hewett, D.~E. Oliver, D.~L. Rubin, K.~L. Easton, J.~M. Stuart, R.~B. Altman, and T.~E. Klein, ``Pharmgkb: the pharmacogenetics and pharmacogenomics knowledge base,'' \emph{Methods in molecular biology}, vol. 1015, pp. 311--320, 2013.

\bibitem{cella2010patient}
D.~Cella, W.~Riley, A.~Stone, N.~Rothrock, B.~Reeve, S.~Yount, D.~Amtmann, R.~Bode, D.~Buysse, S.~Choi \emph{et~al.}, ``The patient-reported outcomes measurement information system (promis): progress of an nih roadmap cooperative group during its first two years,'' \emph{Medical care}, vol.~48, no. 5 Suppl 1, p.~S1, 2010.

\bibitem{smith2007obo}
B.~Smith, M.~Ashburner, C.~Rosse, J.~Bard, W.~Bug, W.~Ceusters, L.~J. Goldberg, K.~Eilbeck, A.~Ireland, C.~J. Mungall \emph{et~al.}, ``The obo foundry: coordinated evolution of ontologies to support biomedical data integration,'' \emph{Nature biotechnology}, vol.~25, no.~11, pp. 1251--1255, 2007.

\bibitem{hripcsak2015observational}
G.~Hripcsak, J.~D. Duke, N.~H. Shah, C.~G. Reich, V.~Huser, M.~J. Schuemie, M.~A. Suchard, R.~W. Park, I.~C. Wong, P.~R. Rijnbeek \emph{et~al.}, ``Observational health data sciences and informatics (ohdsi): opportunities for observational researchers,'' \emph{Studies in health technology and informatics}, vol. 216, p. 574, 2015.

\bibitem{hanna2022chronic}
J.~Hanna, M.~Brochhausen, and W.~R. Hogan, ``Chronic disease ontology toolkit: A structured approach to developing ontologies for chronic disease management,'' \emph{BMJ Health \& Care Informatics}, vol.~29, no.~2, p. e100509, 2022.

\bibitem{michie2017human}
S.~Michie, J.~Thomas, M.~Johnston, P.~M. Aonghusa, J.~Shawe-Taylor, M.~P. Kelly, L.~A. Deleris, A.~N. Finnerty, M.~M. Marques, S.~H. Norgate \emph{et~al.}, ``The human behaviour-change project: harnessing the power of artificial intelligence and machine learning for evidence synthesis and interpretation,'' \emph{Implementation Science}, vol.~12, no.~1, pp. 1--12, 2017.

\bibitem{pistoia_alliance}
\BIBentryALTinterwordspacing
{Pistoia Alliance}, ``Pistoia alliance,'' 2023. [Online]. Available: \url{https://www.pistoiaalliance.org}
\BIBentrySTDinterwordspacing

\bibitem{Millecam2021}
T.~Millecam, A.~J. Jarrett, N.~Young, D.~E. Vanderwall, and D.~Della~Corte, ``Coming of age of allotrope: Proceedings from the fall 2020 allotrope connect,'' \emph{ScienceDirect}, Apr 2021.

\bibitem{VENKATASUBRAMANIAN20061482}
\BIBentryALTinterwordspacing
V.~Venkatasubramanian, C.~Zhao, G.~Joglekar, A.~Jain, L.~Hailemariam, P.~Suresh, P.~Akkisetty, K.~Morris, and G.~Reklaitis, ``Ontological informatics infrastructure for pharmaceutical product development and manufacturing,'' \emph{Computers \& Chemical Engineering}, vol.~30, no.~10, pp. 1482--1496, 2006, papers form Chemical Process Control VII. [Online]. Available: \url{https://www.sciencedirect.com/science/article/pii/S0098135406001578}
\BIBentrySTDinterwordspacing

\bibitem{engproc2021009038}
\BIBentryALTinterwordspacing
E.~Lallas, A.~Karageorgos, and G.~Ntalos, ``An ontology based approach for regulatory compliance of eu reg. no 995/2010 in greece,'' \emph{Engineering Proceedings}, vol.~9, no.~1, 2021. [Online]. Available: \url{https://www.mdpi.com/2673-4591/9/1/38}
\BIBentrySTDinterwordspacing

\bibitem{raebel2013standardizing}
M.~A. Raebel, J.~Schmittdiel, A.~J. Karter, J.~L. Konieczny, and J.~F. Steiner, ``Standardizing terminology and definitions of medication adherence and persistence in research employing electronic databases,'' \emph{Health Services Research}, vol.~48, no.~2, pp. 505--524, 2013.

\bibitem{MORBACH2007147}
J.~Morbach, A.~Yang, and W.~Marquardt, ``Ontocape—a large-scale ontology for chemical process engineering,'' \emph{Engineering Applications of Artificial Intelligence}, vol.~20, no.~2, pp. 147--161, 2007, special Issue on Applications of Artificial Intelligence in Process Systems Engineering.

\bibitem{mann2023susie}
V.~Mann, S.~Viswanath, S.~Vaidyaraman, J.~Balakrishnan, and V.~Venkatasubramanian, ``Susie: Pharmaceutical cmc ontology-based information extraction for drug development using machine learning,'' \emph{Computers \& Chemical Engineering}, vol. 170, p. 108038, 2023.

\bibitem{Arp2015-bk}
R.~Arp, B.~Smith, and A.~D. Spear, \emph{\BIBforeignlanguage{en}{Building Ontologies with Basic Formal Ontology}}, ser. The MIT Press.\hskip 1em plus 0.5em minus 0.4em\relax London, England: MIT Press, Jul. 2015.

\bibitem{qudt}
``qudt.org,'' \url{https://www.qudt.org/pages/QUDToverviewPage.html}.

\bibitem{uom}
``Uom,'' \url{https://units-of-measurement.org/}.

\bibitem{mesh}
``{M}edical {S}ubject {H}eadings,'' \url{https://www.nlm.nih.gov/mesh/meshhome.html}.

\bibitem{ghs}
{United Nations: Economic Commission for Europe}, \emph{\BIBforeignlanguage{en}{Globally harmonized system of classification and labelling of chemicals ({GHS})}}, 10th~ed.\hskip 1em plus 0.5em minus 0.4em\relax New York, NY: United Nations, Jul. 2023.

\bibitem{w3SKOSXL}
``{S}{K}{O}{S} e{X}tension for {L}abels ({S}{K}{O}{S}-{X}{L}),'' \url{https://www.w3.org/TR/skos-reference/skos-xl.html}, 2009.

\bibitem{skohub}
``{S}ko{H}ub,'' \url{https://skohub.io/}.

\bibitem{sparontologiesPatternOntology}
A.~D. Iorio, F.~Vitali, and S.~Peroni, ``{T}he {P}attern {O}ntology,'' \url{https://sparontologies.github.io/po/current/po.html}, 2013.

\bibitem{sparDoCO}
A.~Constantin, S.~Peroni, S.~Pettifer, D.~Shotton, and F.~Vitali, ``The document components ontology ({DoCO}),'' \emph{Semant. Web}, vol.~7, no.~2, pp. 167--181, Feb. 2016.

\bibitem{PERONI201233}
\BIBentryALTinterwordspacing
S.~Peroni and D.~Shotton, ``Fabio and cito: Ontologies for describing bibliographic resources and citations,'' \emph{Journal of Web Semantics}, vol.~17, pp. 33--43, 2012. [Online]. Available: \url{https://www.sciencedirect.com/science/article/pii/S1570826812000790}
\BIBentrySTDinterwordspacing

\bibitem{SigmaAldrich_1003009}
\BIBentryALTinterwordspacing
{Sigma-Aldrich}, ``Acetaminophen (400 mg),'' 2024, accessed: 2024-10-09. [Online]. Available: \url{https://www.sigmaaldrich.com/US/en/sds/usp/1003009?userType=anonymous}
\BIBentrySTDinterwordspacing

\end{thebibliography}
\

\end{document}